\documentclass[10pt,twocolumn,letterpaper]{article}

\usepackage{iccv}
\usepackage{times}
\usepackage{epsfig}
\usepackage{graphicx}
\usepackage{amsmath}
\usepackage{amssymb}
\usepackage[table,xcdraw]{xcolor}
\usepackage{mathtools}
\usepackage{adjustbox}
\usepackage{gensymb}
\usepackage{indentfirst}
\usepackage{multirow}
\usepackage[T1]{fontenc}
\usepackage{mathrsfs}
\usepackage{booktabs}
\usepackage{array}
\usepackage{colortbl}
\usepackage{bm}
\usepackage{lipsum}
\usepackage{subfigure}
\usepackage{color}

\usepackage[pagebackref=true,breaklinks=true,letterpaper=true,colorlinks,bookmarks=false]{hyperref}

\iccvfinalcopy 

\def\iccvPaperID{8332} 

\newcommand{\fig}{Fig.}

\ificcvfinal\fi

\begin{document}

\title{Motion Basis Learning for Unsupervised Deep Homography Estimation\\with Subspace Projection}


\author{
    Nianjin Ye$^{1}$\ \ \ \ Chuan Wang$^{1}$\ \ \ \ Haoqiang Fan$^{1}$\ \ \ \ Shuaicheng Liu$^{2,1}$\thanks {Corresponding author}
  \\
  \\
    $^{1}$Megvii Technology \\ 
    $^{2}$University of Electronic Science and Technology of China \\
}

\maketitle
\ificcvfinal\thispagestyle{empty}\fi

\begin{abstract}
In this paper, we introduce a new framework for unsupervised deep homography estimation. Our contributions are 3 folds. First, unlike previous methods that regress 4 offsets for a homography, we propose a homography flow representation, which can be estimated by a weighted sum of 8 pre-defined homography flow bases. Second, considering a homography contains 8 Degree-of-Freedoms (DOFs) that is much less than the rank of the network features, we propose a Low Rank Representation (LRR) block that reduces the feature rank, so that features corresponding to the dominant motions are retained while others are rejected. Last, we propose a Feature Identity Loss (FIL) to enforce the learned image feature warp-equivariant, meaning that the result should be identical if the order of warp operation and feature extraction is swapped. With this constraint, the unsupervised optimization is achieved more effectively and more stable features are learned. Extensive experiments are conducted to demonstrate the effectiveness of all the newly proposed components, and results show that our approach outperforms the state-of-the-art on the homography benchmark datasets both qualitatively and quantitatively. Code is available at \url{https://github.com/megvii-research/BasesHomo}
\end{abstract}
\vspace{-4mm}

\section{Introduction} \vspace{-1mm}

\begin{figure}[t]
\centering
   \includegraphics[width=1\linewidth]{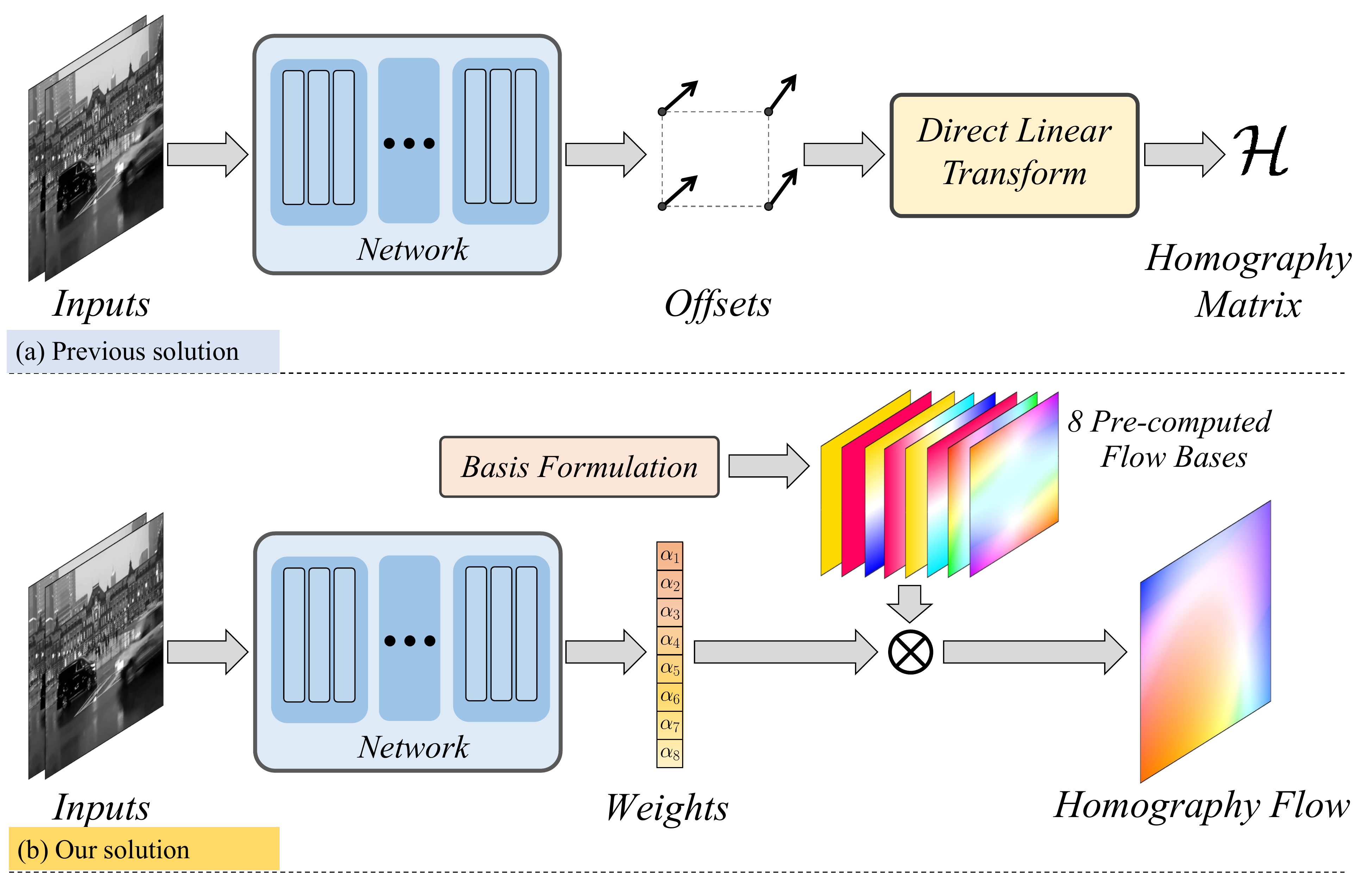}
   \caption{(a) Previous deep homography methods estimate 4 motion offsets and solve a DLT for the result. (b) We construct 8 flow motion bases by modifying matrix elements of a homography, and then regress 8 weights to combine the flow bases for the result.} \vspace{-5mm}
\label{fig:teaser}
\end{figure}

\begin{figure*}[t]
\centering
   \includegraphics[width=0.96\linewidth]{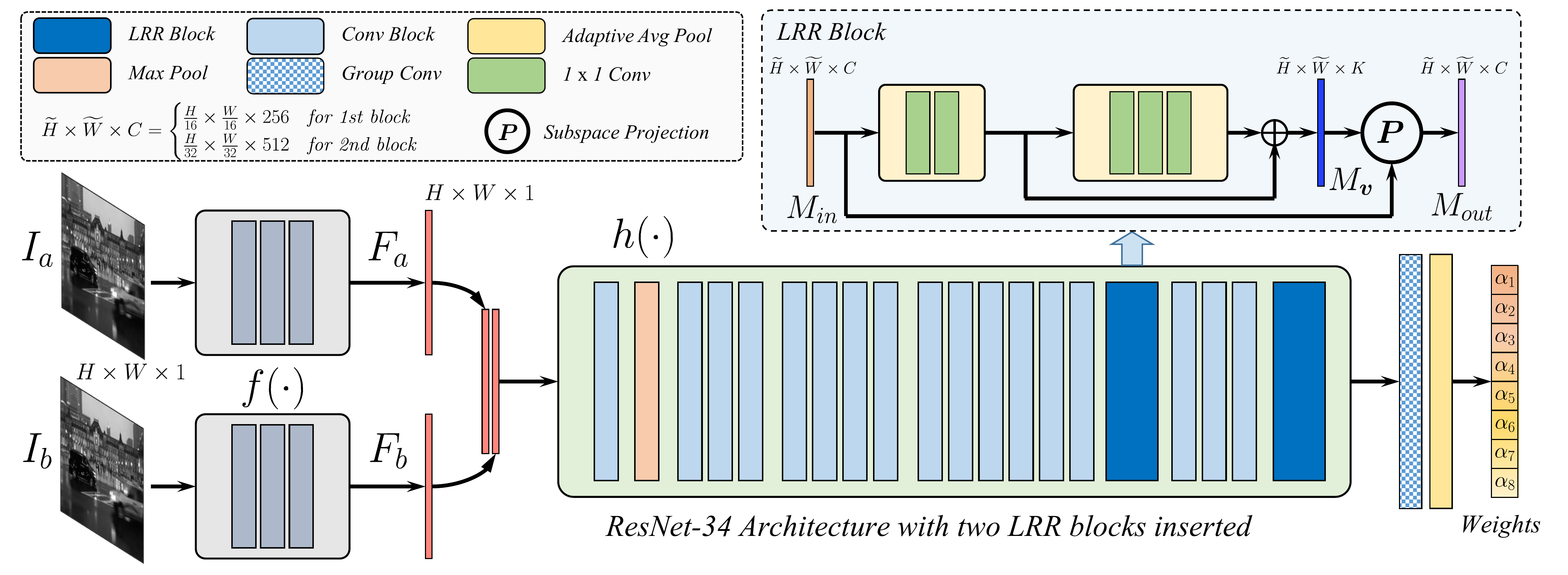}
   \caption{Our network pipeline takes grayscale image patches $I_a$ and $I_b$ as input, and produces 8 weights to combine 8 pre-defined homography bases to produce a homography flow as output. The network consists of two modules, a warp-equivariant feature extractor $f(\cdot)$ and a homography estimator $h(\cdot)$, with 2 inserted LRR blocks to reduce the rank of the motion features. }\vspace{-4mm}
\label{fig:pipeline}
\end{figure*}

Homography is a fundamental and important image alignment model that has been widely used for image registration~\cite{Alex1972Multiple}. A homography is a $3\times3$ matrix that contains 8 Degree-of-Freedoms (DOFs), with each 2 for scale, translation, rotation and perspective~\cite{Hartley2003}. Traditionally, a homography is often estimated by detecting and matching image features~\cite{LoweSIFT04,RubleeRKB11}, and then solving a Direct Linear Transform (DLT)~\cite{Hartley2003} with outlier removal~\cite{FischlerRansac81}. In contrast, deep homography methods take two images as the network input, and directly output a homography matrix~\cite{detone2016deep}. Compared with traditional methods that highly rely on the extracted feature matches, deep methods are more robust. 

Deep methods can be classified into two categories, supervised~\cite{detone2016deep,LeLZA20} and unsupervised~\cite{zhang2020content,NguyenCSTK18}. The former one adopts synthesized examples with ground-truth labels to train the network while the latter one directly minimizes the photometric or feature differences between two images. As synthesized examples cannot reflect scene parallax and dynamic objects, unsupervised methods often generalize better than the supervised ones. For unsupervised methods, Nguyen~\etal~\cite{NguyenCSTK18} minimized error over the entire image while Zhang~\etal~\cite{zhang2020content} proposed to learn a mask to skip outlier regions during the minimization.

It is not optimal to directly regress the elements of a homography matrix, as they are with different magnitudes. Current solution is to regress 4 offsets~\cite{detone2016deep,LeLZA20,zhang2020content,NguyenCSTK18}, which is equivalent to a homography if feeds them to the DLT solver (\fig~\ref{fig:teaser}(a)). In this work, we start from a new direction by proposing a ``\textit{homography flow}'' representation (Fig.~\ref{fig:teaser}(b)). Specifically, we first generate 8 flow bases by modifying the entries of a homography matrix one at a time. As such, 8 homography matrices are obtained, each of which can be further translated into a flow map given the image coordinates, yielding 8 homography flow bases. In \textbf{small-baseline} scenarios, a homography flow can be reconstructed inside the space spanned by these flow bases by learning combinational weights.

As homography has only 8 DOFs, the homography flow lies in a low-rank subspace. However, the rank of the motion features through a network are usually much higher than that of a homography. From this observation, we propose to decrease the rank of the features by projecting them into their subspaces. Specifically, the projection contains two steps, including discovering the subspace bases of the features maps and then transforming feature maps into the subspace. To achieve this projection, we propose a Low Rank Representation (\textbf{LRR}) block, that can be plugged into a normal CNN and be trained end-to-end for the feature rank reduction. When the rank is reduced, features corresponding to the dominant motions, \ie motions that could be described by a homography, are often retained. Features induced from non-dominant motions, e.g., multi-depth and dynamic contents, are often removed or suppressed.

Besides, the triplet loss of previous method still introduces trivial solutions~\cite{zhang2020content}. Specifically, the feature warp-equivariance cannot be well preserved during the unsupervised training, while this property should have held ideally, \ie $f(\mathcal{W}(I)) = \mathcal{W}(f(I))$, where $\mathcal{W}, f(\cdot)$ represent the warp operation and feature extraction. The lack of feature warp-equivariance results in the incorrect optimization of triplet loss, whose convergence direction is dominated by the distances between target features (anchors) and source features (negatives). However, the closer distances between target features (anchors) and warped source features (positives) are more essential regarding the alignment task. To this end, we propose a ``\textit{Feature Identity Loss}'' (\textbf{FIL}) to enforce the image feature to be warp-equivariant. It is demonstrated that with FIL, our model can achieve more effective unsupervised optimization and learn more stable features.


We demonstrate the effectiveness of all the newly proposed techniques and components by extensive experiments and ablation studies. The experimental results also verify that our method outperforms the state-of-the-arts on the public benchmark both qualitatively and quantitatively.
To sum up, our contributions are as follows:
\vspace{-1.5mm}
\begin{itemize}
\setlength{\itemsep}{-1pt}
\setlength{\parsep}{0pt}
\setlength{\parskip}{0pt}
  \item We propose a new representation ``homography flow'' that assembles 8 pre-computed flow bases for unsupervised deep homography estimation.
  \item We propose a new LRR block that reduces motion feature rank so as to reject motion noises implicitly. 
  \item We propose a new FIL loss that enforces the warp-equivariance of the learned image feature to facilitate a stable unsupervised optimization.
\end{itemize}
\vspace{-4mm}


\section{Related works} \vspace{-1mm}
\paragraph{Traditional homography.}
A homography is often estimated by first detecting and matching image features, such as SIFT~\cite{LoweSIFT04}, ORB~\cite{RubleeRKB11}, SURF~\cite{BayTGSURF06}, LPM~\cite{ma2019locality}, GMS~\cite{bian2017gms}, SOSNet~\cite{TianSOSNet19}, LIFT~\cite{YiTLF16}, BEBLID~\cite{suarez2020beblid} and OAN~\cite{zhang2019learning}, after which two sets of point correspondences were established. Next, the false matches were rejected by RANSAC~\cite{FischlerRansac81}, IRLS~\cite{1977Robust}, MAGSAC~\cite{BarathMAGSAC19}. Finally, a DLT is solved for the homography~\cite{daglib_AHAZ}. Some deep approaches have been proposed to improve the feature detection, e.g., SuperPoint~\cite{detone2018superpoint} or matching, e.g., SuperGlue~\cite{sarlin2020superglue}. 

\vspace{-5mm}
\paragraph{Deep homography.}
The deep homography can be classified into supervised~\cite{detone2016deep, LeLZA20} and unsupervised~\cite{NguyenCSTK18,zhang2020content} methods.
Compared to supervised approaches which learn transformation from synthesized images that lack depth disparities, the unsupervised ones can be trained on real image pairs by minimizing the photometric loss between two images, with Spatial Transform Network (STN)~\cite{jaderberg2015spatial} to warp the source to the target.
Nguyen~\emph{et al.}~\cite{NguyenCSTK18} minimizes the photometric loss on the entire image while Zhang~\emph{et al.}~\cite{zhang2020content} learns a mask to skip outlier regions. 

\vspace{-5mm}
\paragraph{Bases learning.}
Our method is also related to bases learning~\cite{liu2012robust}. 
Tang~\emph{et al.} shows that there are subspaces that can be exploited for regularization in low-level vision problems~\cite{tang2020lsm}. PCAFlow learns flow bases from movies, showing that the flow estimation can be converted into the learning of the weighted sum of the learned flow bases~\cite{wulff2015efficient}. Inspired by these works, we learn the coefficients to combine 8 pre-defined flow bases to estimate a homography flow.


\vspace{-2mm} 

\section{Algorithm}\label{sec:method} \vspace{-1mm}

\subsection{Network Structure}\vspace{-1mm}
Our method is built on convolutional neural networks, which takes two gray image patches $I_a$ and $I_b$ of size $H\times W$ as input, and produces a homography flow $\bm{H}_{ab}$ from $I_a$ to $I_b$ of the same size as output. The entire structure consists of two modules, a warp-equivariant feature extractor $f(\cdot)$ and a homography estimator $h(\cdot)$. $f(\cdot)$ is a fully convolutional network which accepts input of arbitrary sizes, and $h(\cdot)$ adapts a backbone of ResNet-34~\cite{he2016identity} with our newly introduced LRR blocks and produces 8 weights, which are used to linearly combine 8 pre-computed flow bases to obtain $\bm{H}_{ab}$. \fig~\ref{fig:pipeline} illustrates the network structure. 

\begin{figure}[t]
\centering
   \includegraphics[width=0.8\linewidth]{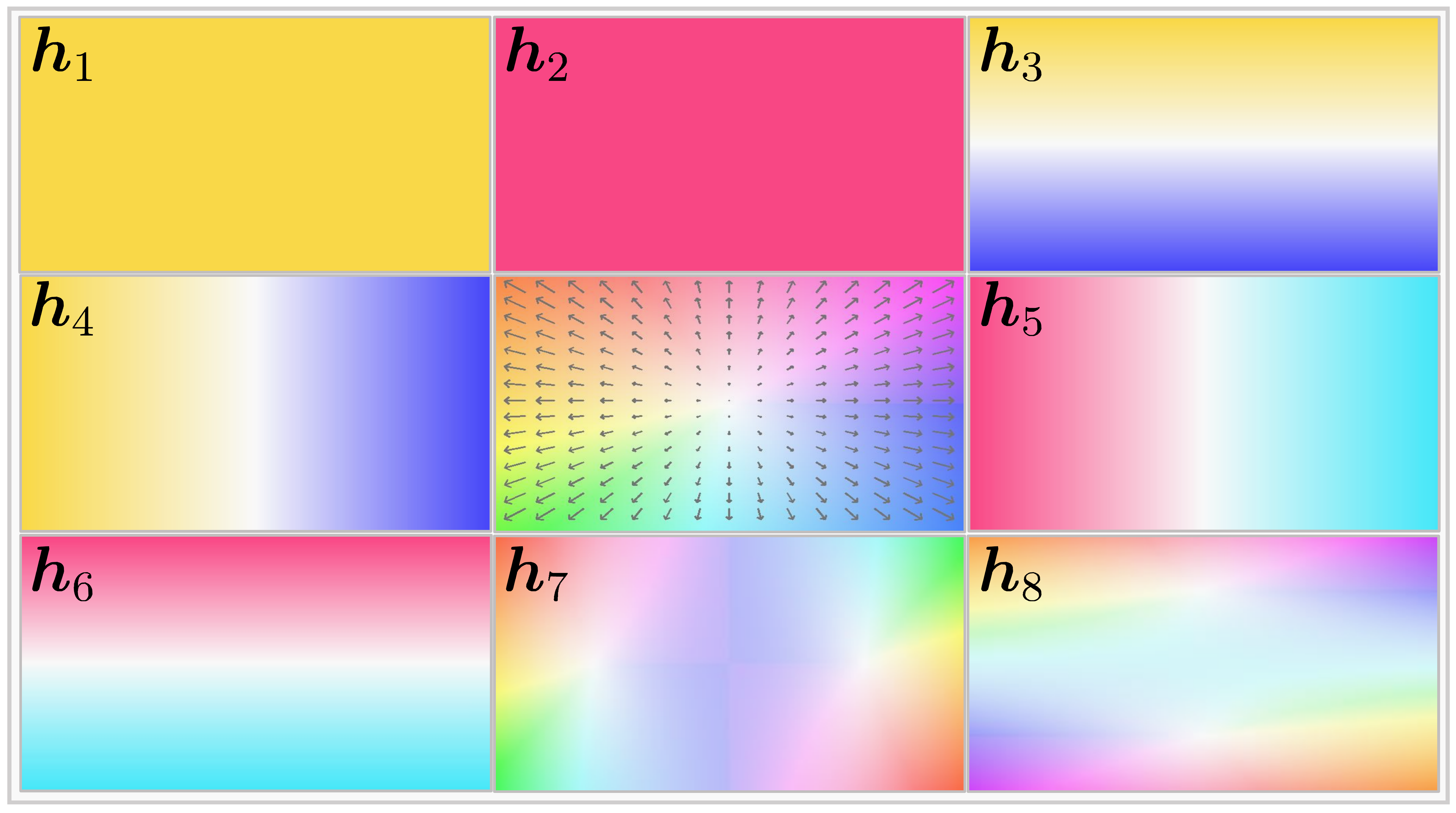}
   \caption{The visualization of the pre-computed 8 orthogonal and normalized flow bases $\bm{h}_1\sim\bm{h}_8$ and the flow legend in the center.}\vspace{-5mm}
\label{fig:basis}
\end{figure}

\vspace{-3mm}\paragraph{Homography flow and its basis formulation.}\label{sec:homo-flow-basis-formulation}
A homography matrix has 8 DOFs and it is computed by solving a DLT after 4 corner offsets of an image are predicted as in~\cite{zhang2020content,NguyenCSTK18}. In this paper, we solve the problem from a new perspective. Specifically, our network learns a special optical flow of size $H\times W\times 2$ constrained by the homography, called ``\textit{homography flow}''. Due to the constraint, homography flow falls into a 8-D subspace within the entire $2HW$-D space of a general optical flow. It can be represented by 8 orthogonal flow bases spanning the subspace, \ie
\begin{equation}\label{eq:8-bases-exist}
{
\begin{aligned}
\exists~\{\bm{h}_i\}~~\text{s.t.}~~\bm{h}_{ab} &= \sum_i \alpha_i \bm{h}_i~~(i = 1,2,...,8) \\
\text{where}~\bm{h}_i &\in \mathbb{R}^{2HW},~\bm{h}_i^T\bm{h}_j = 0
\end{aligned}}
\end{equation}
Here $\bm{h}_{ab}$ is the flattened version of $\bm{H}_{ab}$, and $\{\alpha_i\}$ are the coefficients for the flow bases.

To obtain the orthogonal flow bases, we first generate 8 homography matrices by modifying each single entry $h_i$ of an identity homography matrix, except the entry at the position of $(3,3)$ which is always normalized to 1. Given the image coordinates, a homography matrix can be converted to a flow map by transforming the image coordinates and minus their original positions. Then, 8 homography flows are normalized by their  largest flow magnitude followed by a QR decomposition. Mathematically, it is described as,
\begin{align}\label{eq:QR-decompose}
\bm{M} = \bm{Q}\cdot\bm{R}\quad(\bm{M},\bm{Q} \in \mathbb{R}^{2HW\times 8}, \bm{R}\in\mathbb{R}^{8\times8})
\end{align}
where each column of matrix $\bm{M}$ is the flattened normalized homography flow $\bm{H}_i$ as mentioned above. By QR decomposition, columns of $\bm{Q}$ are orthogonal and they naturally serve as the flow bases spanning the homography subspace, \ie $\bm{Q} = [\bm{h}_1,\bm{h}_2,...,\bm{h}_8]$. In other words, each flow basis is associated with a tangent space at the origin of the homography group. With the 8 bases formulated, a homography flow can be acquired by accurately predicting their weights $\{\alpha_i\}$. Considering the perspective transformation can be approximated well with linear model in small baseline tasks, we can use such a linear-weighted solution to approximate homography. We visualize the bases in \fig~\ref{fig:basis}.

\textbf{Discussion.} Compared with the aforementioned bases learning methods like PCAFlow~\cite{wulff2015efficient}, our method has similarity to them so that a set of homography bases could be potentially learned. However, our solution has its specificity because unlike PCAFlow~\cite{wulff2015efficient} which predicts a general optical flow that requires more flexibility, we only deal with the background of \textit{small-baseline} scenes. It means that less flexibility exists in the solution space so that an analytical derivation of ``homography flow'' becomes feasible and complicated learning tools like PCA~\cite{pearson1901pca} could be unemployed. So for simplicity, we just employ the pre-computed bases in this work, although there indeed exist types of homographies that cannot be represented precisely by them, such as those in large-baseline scenes. 



\vspace{-3mm}\paragraph{Warp-equivariant feature extractor.}
Before~\cite{zhang2020content}, previous unsupervised DNN based methods commonly minimize the pixel intensity values for the registration. In~\cite{zhang2020content}, the authors proposed to minimize the difference of learned deep features instead of using the original images. In this paper, we similarly follow the idea of~\cite{zhang2020content}, but constrain the learned features with warp-equivariance, which means the results should be identical if we swap the order of warp operation $\mathcal{W}$ and feature extraction $f$ given an input image $I$, \ie $\mathcal{W}\big(f(I)\big) = f\big(\mathcal{W}(I)\big)$. For inputs $I_a$ and $I_b$, the feature extractor shares weights and produces feature maps $F_a$ and $F_b$. In practice, features with absolute warp-equivariance are rarely achieved. Thus we introduce a new loss $\mathbf{L}_{\mathcal{W}} = \big|\mathcal{W}\big(f(I)\big) - f\big(\mathcal{W}(I)\big)\big|$ as a constraint to approximate this property, which is detailed in Sec.~\ref{sec:warp-equivariance}.

\vspace{-3mm}\paragraph{Homography estimator with LRR blocks.}
Given the feature maps $F_a$ and $F_b$, we concatenate them to build a feature map $[F_a, F_b]$. Then, it is fed to the homography estimator network to produce 8 weights. These weights linearly combine $\{\bm{h}_i\}$ to produce the final homography flow $\bm{H}_{ab}$. We use $h(\cdot)$ to represent the whole process, \ie
\begin{align}\label{eq:homo-estimator}
\bm{H}_{ab} = h([F_a, F_b])
\end{align}\vspace{-3mm}

The backbone of $h(\cdot)$ generally follows a ResNet-34~\cite{he2016identity} structure, except that two newly introduced LRR blocks are inserted at two layers. Each LRR block consists of shallow residual convolution layers and learns $K$ bases for the input motion feature forwarded by the preceding layers. It then generates an output motion feature of rank at most $K$ by subspace projection. Specifically, given an input motion feature $M_{in} \in \mathbb{R}^{H\times W\times C}$, the residual convolution layers convert it into a feature ${M}_{\bm{v}} \in \mathbb{R}^{H\times W\times K}$ of $K$ channels. Then each channel serves as a feature basis $\bm{v}_k \in \mathbb{R}^{HW}, k=1,2,...,K$ after being flattened. Finally, we project $M_{in}$ into the subspace of the feature bases to obtain a low-rank motion feature $M_{out} \in \mathbb{R}^{H\times W\times C}$, \ie
\begin{align}
M_{out} = \bm{V}(\bm{V}^T\bm{V})^{-1}\bm{V}^T\cdot M_{in}
\end{align}
where $\bm{V} = [\bm{v}_1,\bm{v}_2,...,\bm{v}_K] \in \mathbb{R}^{HW\times K}$. Note that the normalization term $(\bm{V}^T\bm{V})^{-1}$ is required since the feature bases $\{\bm{v}_k\}$ are not ensured orthogonal. 

After the 2nd LRR block, the motion feature computed is forwarded to a group convolution and an adaptive pooling layer, to produce the final 8 weights $\{\alpha_i\}$ for homography flow bases combination. We illustrate the structure of LRR block in \fig~\ref{fig:pipeline}, and $K$ is set to 16 in all of our experiments.

\begin{figure}
  \centering
  \includegraphics[trim=3.cm 0.9cm 3.8cm 1cm,clip,width=0.98\linewidth]{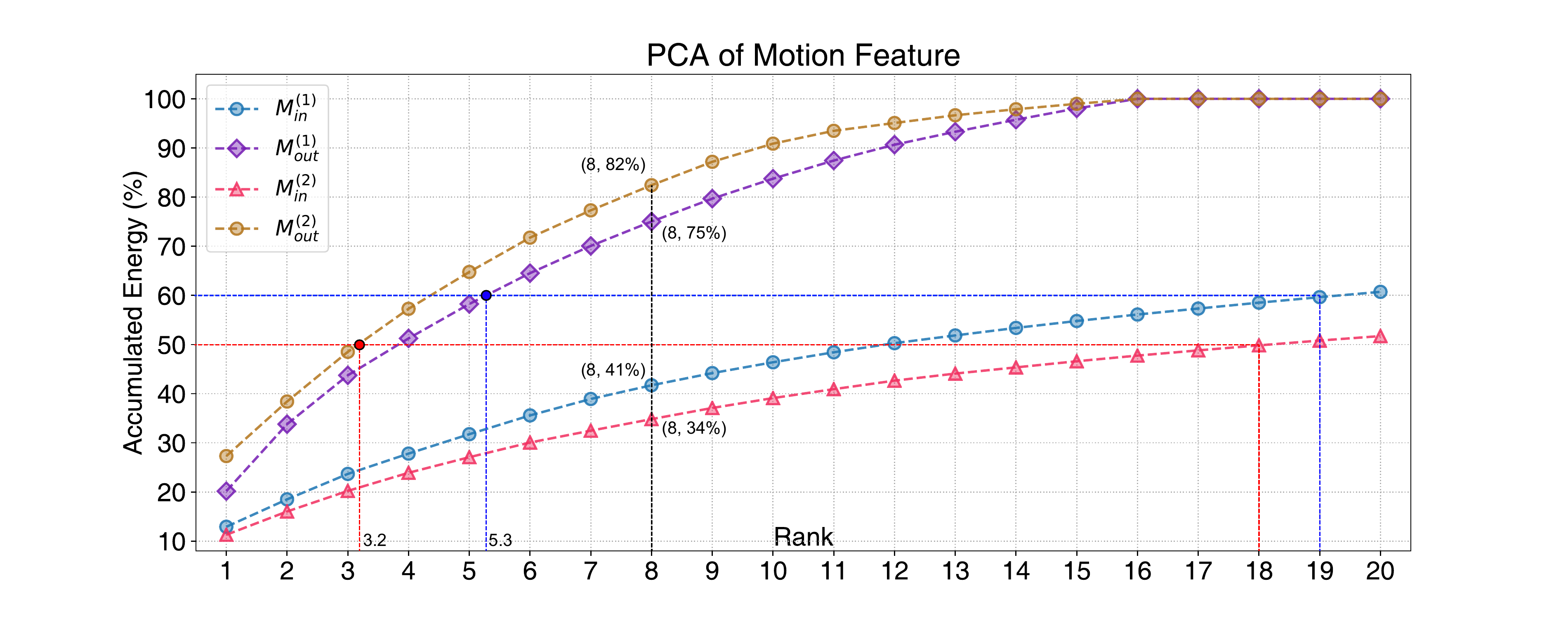}\\
  \caption{Accumulated energy of the principal components of the motion features, before and after the 1st and 2nd LRR blocks.}\label{fig:motion-feature-energy}\vspace{-3mm}
\end{figure}
\subsection{Robust Homography Estimation by LRR}\label{sec:lrr}\vspace{-1mm}
As indicated in Sec.~\ref{sec:homo-flow-basis-formulation}, a homography flow is of low rank, which means the rank of the motion features through various layers in $h(\cdot)$ should be reduced. Our observation is that, if the rank of the motion feature is reduced, the latent weights for the homography flow bases could be predicted more accurately and easily, where motion outliers are excluded during the rank reduction.

The motion outliers are motions caused by dynamic contents or non-planar depth variations outside the solution space of a single homography. Traditionally, motion outliers are often rejected by RANSAC~\cite{LoweSIFT04}. In DNN, Zhang~\etal's~\cite{zhang2020content} predicted a mask to skip the motion outliers. In this paper, LRR block serves this purpose. It reduces the rank of the motion feature, during which the feature ranks corresponding to the outliers are reduced. Because we enforce the network to learn motions spanned by the homography bases, any motions outside the subspace are treated as motion noise. In this way, the non-homography motion can be eliminated automatically during the rank reduction. As a result, a mask predictor as is no longer needed~\cite{zhang2020content}.

As seen in Table~\ref{tab:compare}, with LRR block inserted, our network produces a lower error and performs much better compared to~\cite{zhang2020content}. We also analyze the accumulated energy of the principal components for the motion feature $M_{in}$ and $M_{out}$. As seen in \fig~\ref{fig:motion-feature-energy}, after the 1st LRR block, the number of principal components (NPC) are reduced from 19 to 5.3 in terms of accumulated energy being 60\%. The effectiveness of the 2nd LRR block is more obvious, reducing the NPC from 18 to 3.2 in terms of 50\% energy, reflecting the rank of the motion feature is highly reduced. 


\begin{figure}[t]
\centering
   \includegraphics[width=1\linewidth]{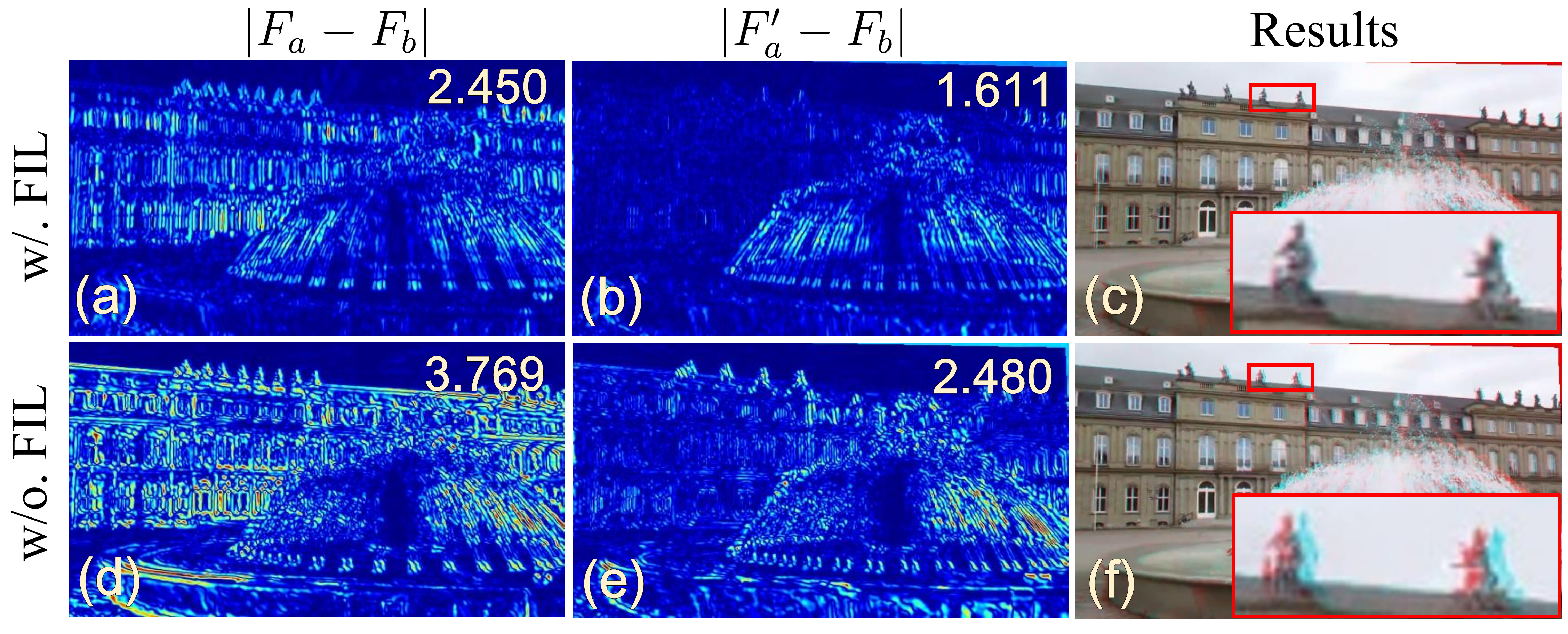}
   \caption{The comparison between with and without the warp-equivariance by FIL. Please refer the text for more details.} \vspace{-6mm}
\label{fig:FIL}
\end{figure}

\begin{figure*}[t]
\centering
   \includegraphics[width=0.95\linewidth]{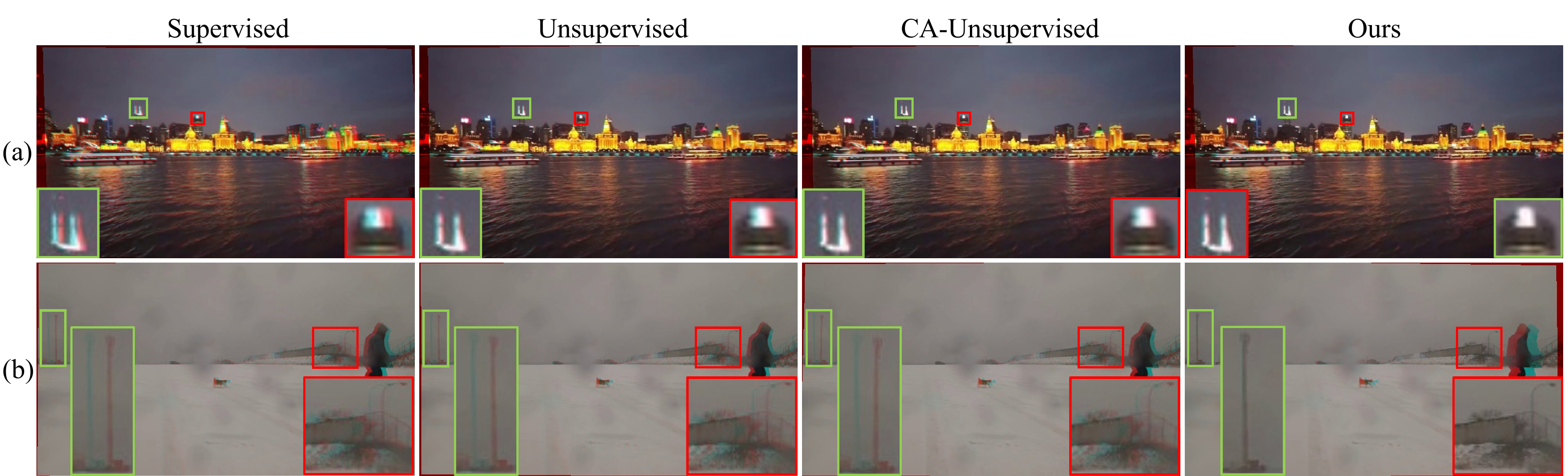}
   \caption{Qualitative comparison with recent DNN-based approaches. Columns 1 $\sim$ 4 are results of supervised~\cite{detone2016deep}, Unsupervised~\cite{NguyenCSTK18}, CA-Unsupervised~\cite{zhang2020content} and ours. The alignment difficulty of Rows 2 is greater than Row 1.}\vspace{-4mm}
\label{fig:DNNbased}
\end{figure*}

\subsection{Triplet Loss with Feature Warp-Equivariance}\label{sec:warp-equivariance}\vspace{-1mm}
With the homography flow $\bm{H}_{ab}$ estimated, we warp feature map $F_a$ to $F'_a$ and formulate a triplet loss without involving an attention mask as in~\cite{zhang2020content}, \ie
\begin{equation}\label{eq:triplet-loss}
\small{\mathbf{L}_{\mathcal{T}}(I_a, I_b) = \mathbf{L}_{\mathcal{T}}^{ab} = |F'_a - F_b|_1 - |F_a - F_b|_1}
\end{equation}

The original idea on the triplet loss tries to learn a discriminative feature and an accurate homography simultaneously to well align the input images. Even though it has been demonstrated successful in most cases in~\cite{zhang2020content}, it still has the potential to be incorrectly optimized so that $|F_a - F_b|$ is over-maximized while $|F'_a - F_b|$ is still under-minimized, due to the enough DOFs of $f(\cdot)$. To this end, we add a new constraint named ``\textit{Feature Identity Loss}'' (FIL) to preserve the warp-equivariance of the learned feature, meaning that the final feature should be approximately identical if we swap the order of warp operation and feature extraction, \ie
\begin{equation}\label{eq:FIL}
\small{\mathbf{L}_{\mathcal{W}}(I_a, f, \mathcal{W}_{ab}) = \mathbf{L}^{ab}_{\mathcal{W}} = |\mathcal{W}_{ab}(f(I_a)) - f(\mathcal{W}_{ab}(I_a))|_1}
\end{equation}
where $\mathcal{W}_{ab}$ is the warp operation by homography flow $\bm{H}_{ab}$. 

We observe that with this constraint, the optimization of $f(\cdot)$ can be stabilized, improving the accuracy of estimated homography flow. We visualize an example in \fig~\ref{fig:FIL}, where without FIL involved, the triplet loss $\mathbf{L}_{\mathcal{T}}(I_a, I_b)$ is less than the one optimized with FIL, even though $|F'_a - F_b|$ becomes larger. The reason behind is that the term $|F_a - F_b|$ is maximized excessively without FIL. As a result, the alignment accuracy is down-graded. 

In practice, we also swap the order of $I_a$ and $I_b$ to compute symmetric losses $\mathbf{L}(I_b, I_a)$ and $\mathbf{L}_{\mathcal{W}}(I_b, f, \mathcal{W}_{ab})$, and add a constraint to enforce the homograph flows $\bm{H}_{ba}$ and $\bm{H}_{ab}$ to be inverse. So, the energy is written as follows,
\begin{equation}\label{eq:optimization}
\small{\min_{f, h}~(\mathbf{L}_{\mathcal{T}}^{ab} + \mathbf{L}_{\mathcal{T}}^{ba}) + \lambda(\mathbf{L}_{\mathcal{W}}^{ab} + \mathbf{L}_{\mathcal{W}}^{ba}) + \mu|\bm{H}_{ab} + \bm{H}_{ba}|_2^2}
\end{equation}
where $\lambda$ and $\mu$ are set to 1.0 and 0.001 in our experiments. 

\vspace{-2mm}

\section{Experiment}


\begin{figure*}[t]
\centering
   \includegraphics[width=0.97\linewidth]{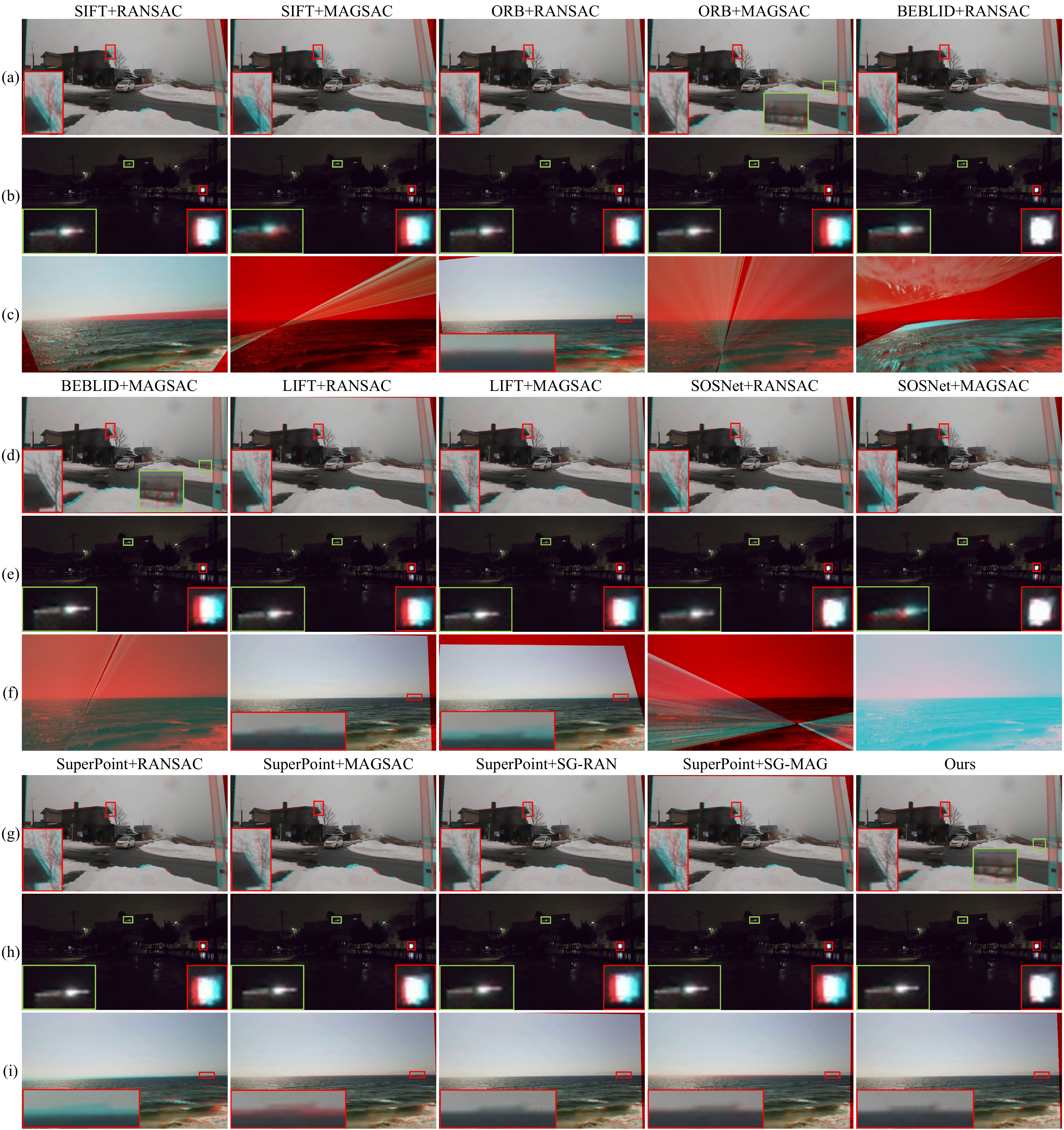}
   \caption{Qualitative comparison with with Feature-based approaches on 3 examples. The combination of various descriptors and outlier elimination methods produced a total of 14 approaches.}\vspace{-4mm}
\label{fig:Featurebased}
\end{figure*}

\begin{figure}[t]
\centering
   \includegraphics[trim=3cm 0.5cm 3.4cm 1cm,clip,width=1\linewidth]{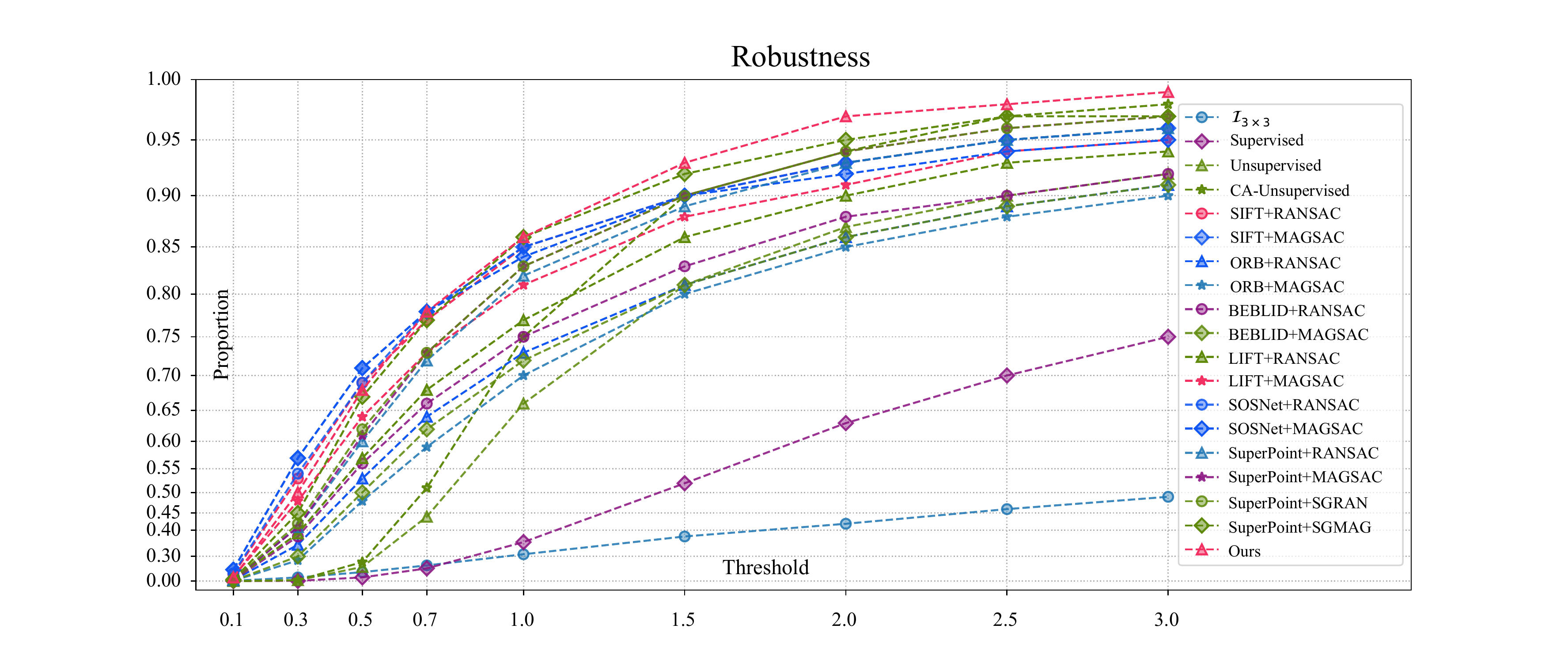}
   \caption{The proportion of inliers in all matching point pairs under various thresholds for all methods. Higher position of curves represent higher robustness.}
\label{fig:robust}\vspace{-4mm}
\end{figure}

\subsection{Dataset and Implementation Details}\vspace{-1mm}
We evaluate our method using the same dataset as in Zhang~\etal's~\cite{zhang2020content}, \ie the CA-Unsupervised. The training set consists of 5 categories of small-baseline image pairs in real scenes, including regular (RE), low-texture (LT), low-light (LL), small-foregrounds (SF), and large-foregrounds (LF). Except for the RE, the other 4 scenes are challenging for the homography estimation. 
A randomly selected subset of $4.2k$ samples is used as the test set, each sample of which contains 6 pairs of labeled matching points for evaluation.

Our network is trained with $360k$ iterations by the Adam optimizer~\cite{KingmaB14}, with parameters $l_r = 10^{-4}$, $\beta_1=0.9$, $\beta_2=0.999$, $\epsilon=10^{-8}$. The batch size is set to 16 and the $l_r$ is reduced by $20\%$ every epoch. 
The implementation is based on PyTorch and the network training is performed on one NVIDIA RTX 2080 Ti. To alleviate the impact of empty boundaries in the warped image, we randomly crop patches of size $320 \times 576$ from the original images to serve as input. \vspace{-4mm} 
\begin{table*}[t]
  \centering
  \small
  \resizebox{0.98\linewidth}{!}{
    \begin{tabular}{
    r
    >{\arraybackslash}p{4.4cm}
    >{\centering\arraybackslash}p{2.2cm}
    >{\centering\arraybackslash}p{2.2cm}
    >{\centering\arraybackslash}p{2.2cm}
    >{\centering\arraybackslash}p{2.2cm}
    >{\centering\arraybackslash}p{2.2cm}
    >{\centering\arraybackslash}p{2.2cm}}
    \toprule
     1) &       & RE    & LT     & LL    & SF    & LF    & Avg \\
    \midrule
     2) & $\mathcal{I}_{3\times3}$ & 7.75(+2483.33\%) & 7.65(+868.35\%) & 7.21(+930.00\%) & 7.53(+960.56\%) & 3.39(+621.28\%) & 6.70(+963.49\%) \\
    \midrule
     3) & Supervised~\cite{detone2016deep} & 1.51(+403.33\%) & 4.48(+467.09\%) & 2.76(+294.29\%) & 2.62(+269.01\%) & 3.00(+538.30\%) & 2.87(+355.56\%) \\
     4) & Unsupervised~\cite{NguyenCSTK18}& 0.79(+163.33\%)& 2.45(+210.13\%)& 1.48(+111.43\%)& 1.11(+56.34\%)& 1.10(+134.04\%)& 1.39(+120.63\%)\\
     5) & CA-Unsupervised~\cite{zhang2020content} & 0.73(+143.33\%)& 1.01(+27.85\%)& 1.03(+47.14\%)& 0.92(+29.58\%)& 0.70(+48.94\%)& 0.88(+39.68\%)\\
     \midrule
     6) & SIFT~\cite{LoweSIFT04} + RANSAC~\cite{FischlerRansac81} & \textcolor{blue}{0.30(+0.00\%)} & 1.34(+69.62\%)& 4.03(+475.71\%)& 0.81(+14.08\%)& 0.57(+21.28\%)& 1.41(+123.81\%)\\
     7) & SIFT~\cite{LoweSIFT04} + MAGSAC~\cite{BarathMAGSAC19} & 0.31(+3.33\%)& 1.72(+117.72\%)& 3.39(+384.29\%)& 0.80(+14.08\%)& \textcolor{blue}{0.47(+0.00\%)} & 1.34(+112.70\%)\\
     8) & ORB~\cite{RubleeRKB11} + RANSAC~\cite{FischlerRansac81} & 0.85(+183.33\%)& 2.59(+227.85\%)& 1.67(+138.57\%)& 1.10(+54.03\%)& 1.24(+163.83\%)& 1.48(+134.92\%)\\
     9) & ORB~\cite{RubleeRKB11} + MAGSAC~\cite{BarathMAGSAC19} & 0.97(+223.33\%)& 3.34(+322.78\%)& 1.58(+125.71\%)& 1.15(+61.97\%)& 1.40(+197.87\%)& 1.69(+168.25\%)\\
    10) & BEBLID~\cite{suarez2020beblid} + RANSAC~\cite{FischlerRansac81} & 0.78(+160.00\%)& 2.83(+258.23\%)& 1.38(+97.14\%)& 1.04(+46.48\%)& 1.33(+182.98\%)& 1.47(+133.33\%)\\
    11) & BEBLID~\cite{suarez2020beblid} + MAGSAC~\cite{BarathMAGSAC19} & 0.94(+213.33\%)& 3.73(+372.15\%)& 3.49(+398.57\%)& 1.17(+64.79\%)& 1.25(+165.96\%)& 2.12(+236.51\%)\\
    \midrule
    12) & LIFT~\cite{YiTLF16} + RANSAC~\cite{FischlerRansac81} & 0.40(+33.33\%)& 2.01(+154.43\%)& 1.14(+62.86\%)& 0.77(+8.45\%)& 0.68(+44.68\%)& 1.00(+58.73\%)\\
    13) & LIFT~\cite{YiTLF16} + MAGSAC~\cite{BarathMAGSAC19} & 0.35(+16.67\%)& 1.85(+134.18\%)& 0.96(+37.14\%)& 0.72(+1.41\%) & 0.50(+6.38\%)& 0.88(+39.68\%)\\
    14) & SOSNet~\cite{TianSOSNet19} + RANSAC~\cite{FischlerRansac81} & \textcolor{red}{0.29(-3.33\%)} & 2.42(+206.33\%)& 3.71(+430.00\%)& 0.77(+8.45\%)& 0.59(+25.53\%)& 1.56(+147.62\%)\\
    15) & SOSNet~\cite{TianSOSNet19} + MAGSAC~\cite{BarathMAGSAC19} & \textcolor{blue}{0.30(+0.00\%)}& 3.00(+279.75\%)& 3.66(+422.86\%)& 0.87(+22.54\%)& 0.49(+4.26\%)& 1.67(+165.08\%)\\
    16) & SuperPoint~\cite{detone2018superpoint} + RANSAC~\cite{FischlerRansac81} & 0.43(+43.33\%)& 0.85(+7.59\%)& 0.77(+10.00\%)& 0.84(+18.31\%)& 0.80(+70.21\%)& 0.74(+17.46\%)\\
    17) & SuperPoint~\cite{detone2018superpoint} + MAGSAC~\cite{BarathMAGSAC19} & 0.45(+50.00\%)& 0.90(+13.92\%)& 0.77(+10.00\%)& 0.76(+7.04\%)& 0.67(+42.55\%)& 0.71(+12.70\%)\\
    18) & SuperPoint~\cite{detone2018superpoint} + SG-RAN~\cite{sarlin2020superglue}~\cite{FischlerRansac81} & 0.41(+36.67\%)& 0.87(+10.13\%)& 0.72(+2.86\%)& 0.80(+12.68\%)& 0.75(+59.57\%)& 0.71(+12.70\%)\\
    19) & SuperPoint~\cite{detone2018superpoint} + SG-MAG~\cite{sarlin2020superglue}~\cite{BarathMAGSAC19} & 0.36(+20.00\%)& \textcolor{blue}{0.79(+0.00\%)} & \textcolor{blue}{0.70(+0.00\%)} & \textcolor{blue}{ 0.71(+0.00\%)} & 0.70(+48.94\%)& \textcolor{blue}{0.63(+0.00\%)} \\
    \midrule
    20) & Ours & \textcolor{red}{0.29(-3.33\%)}& \textcolor{red}{0.54(-31.65\%)} & \textcolor{red}{0.65(-7.14\%)} & \textcolor{red}{0.61(-14.08\%)} & \textcolor{red}{0.41(-12.77\%)} & \textcolor{red}{0.50(-20.63\%)} \\
    \bottomrule
    \end{tabular}
    }\vspace{1mm}
    \caption{Comparison of point matching errors between ours and all other methods. SG-RAN and SG-MAG are SuperGlue~\cite{sarlin2020superglue} + RANSAC~\cite{FischlerRansac81} and SuperGlue~\cite{sarlin2020superglue} + MAGSAC~\cite{BarathMAGSAC19} respectively. The percentage in the bracket indicates the improvements over the second best results. \textcolor{red}{Red} indicates the best performance and \textcolor{blue}{Blue} refers to the second best result.}

    \label{tab:compare}\vspace{-3mm}
\end{table*}%

\subsection{Comparison to Existing Methods}\vspace{-1mm}
We compare our method with two groups of homography estimation approaches, the DNN-based ones and the feature-based ones. The former group includes Supervised~\cite{detone2016deep}, Unsupervised~\cite{NguyenCSTK18} and CA-Unsupervised~\cite{zhang2020content}, and the latter group includes 14 methods, including 12 combinations of 6 types of features (3 traditional: SIFT~\cite{LoweSIFT04} / ORB~\cite{RubleeRKB11} / BEBLID~\cite{suarez2020beblid} and 3 DNN-based: LIFT~\cite{YiTLF16} / SOSNet~\cite{TianSOSNet19} / SuperPoint~\cite{detone2018superpoint}) and 2 outlier rejection algorithms (RANSAC~\cite{FischlerRansac81} / MAGSAC~\cite{BarathMAGSAC19}), as well as 2 additional customized descriptor matching approaches SuperGlue~\cite{sarlin2020superglue} specifically for SuperPoint~\cite{detone2018superpoint} only.

\paragraph{Qualitative comparison.}\vspace{-4mm}
Fig.~\ref{fig:DNNbased} shows the comparison with DNN-based methods. Fig.~\ref{fig:DNNbased}(a) is from the LL category with repetitive dynamic textures at the river, where the Supervised~\cite{detone2016deep} fails because it is trained on synthetic data without dynamic contents. The results of the Unsupervised~\cite{detone2016deep} and the CA-Unsupervised~\cite{zhang2020content} contain some small errors since both methods cannot reject the dynamic flowing rivers precisely. In contrast, our outlier rejection is achieved via LRR, obtaining accurate principal feature attention compared to the mask of CA-Unsupervised~\cite{zhang2020content}.

Our method shows superiority in Fig.~\ref{fig:DNNbased}(b). This scene is from the LT category, where texture quality is extremely poor as in the snow and the sky. The walking man in the foreground making the task much more challenging. Other methods try to align the moving man since he has more textures than his surrounding areas, while only our method successfully aligns the scene without paying attention to the man, demonstrated by the highlighted pole region.


Fig.~\ref{fig:Featurebased} further compares our method with all the aforementioned feature-based solutions. In Fig.~\ref{fig:Featurebased}(a), (d) and (g), we validate various feature-based methods in a snow scene, where all of the feature-based methods fail to produce satisfying results, due to either the feature extraction or the foreground interferences. In contrast, our method aligns this scene more accurately.
As for the latter two examples including a low-light and a low-texture scene, both of them challenge the feature detection and matching. For instance, in the low-light scene (Fig.~\ref{fig:Featurebased}(b), (e) and (h)), only a small portion of the image contains salient regions. In the sea example (Fig.~\ref{fig:Featurebased}(c), (f) and (i)), it is tough to obtain reliable feature matches on the sea textures. In contrast, our method is naturally more adaptable to the case of insufficient features, benefiting from the pursuit of low-rank features.

\paragraph{Quantitative comparison.}\vspace{-5mm}
As seen in Table~\ref{tab:compare}, for each pair of test images, the average $l_2$ distances between the warped source points and the target points are considered as the error metric. We report the errors with respect to each scene category. Specifically, Row $3\sim5$ are deep homography methods; Row $6\sim11$ are traditional feature-based methods and Row $12\sim19$ are DNN-based feature methods. $\mathcal{I}_{3\times3}$ refers to an identity homography, which reflects the original distance between point pairs. 



As for the RE scene, the abundant texture provides sufficient high-quality features for homography estimation. So that the feature-based solutions show obvious advantages in this category. Nevertheless, our method and the combination of SOSNet~\cite{TianSOSNet19} and RANSAC~\cite{FischlerRansac81} outperform the others and achieve the lowest error of 0.29.

For the scenes of LT and LL, most of the feature-based solutions become less robust due to the difficulty in extracting effective features. In contrast, our method consistently works well. In particular, the results of the 2nd best method which is constituted by the 3 latest algorithms (SuperPoint~\cite{detone2018superpoint}, SuperGlue~\cite{sarlin2020superglue} and MAGSAC~\cite{BarathMAGSAC19}) achieve strong performance. In the scenes containing small (SF) and large (LF) foreground, although sufficient texture features are available, dynamic objects and multi-plane occlusions cause troubles for the outlier removal. In our method, objects and multi-plane depth tend to introduce high-rank features in the encoding space, which are abandoned by our low-rank property of LRR blocks. Therefore, our method achieves at least $14.08\%$ and $12.77\%$ lower errors than others on SF and LF, respectively. On the whole, the combination of SuperPoint~\cite{detone2018superpoint}, SuperGlue~\cite{sarlin2020superglue} and MAGSAC~\cite{BarathMAGSAC19} produces rather competitive results for all the scenes, but their average errors are still higher than ours by $20.63\%$.

\paragraph{Robustness evaluation.} \vspace{-5mm}
Furthermore, we evaluate the robustness of all the methods by setting a threshold to judge if a homography matches the marked points. Points with errors lower than the threshold are considered matched inliers, otherwise they are judged as outliers. As such, we report a percentage of matched points overall marked points given a homography estimation method and a threshold, so that a series of curves are reported in \fig~\ref{fig:robust} by setting threshold to 0.1 to 3.0. As seen, our method (the red curve) achieves the highest inlier percentage if the threshold is greater than $0.8$, indicating our method can handle intractable cases better than the others. It draws a similar conclusion as Table~\ref{tab:compare} that our method outperforms others in challenging scenarios, such as LT and LF scenes.
\begin{table*}[t]
  \centering
  \small
  \resizebox{0.98\linewidth}{!}{
    \begin{tabular}{
    r
    >{\centering\arraybackslash}p{1.5cm}
    >{\centering\arraybackslash}p{1.5cm}
    >{\centering\arraybackslash}p{1.0cm}
    >{\centering\arraybackslash}p{2cm}
    >{\centering\arraybackslash}p{2cm}
    >{\centering\arraybackslash}p{2cm}
    >{\centering\arraybackslash}p{2cm}
    >{\centering\arraybackslash}p{2cm}
    >{\centering\arraybackslash}p{2cm}}
    \toprule
     1) & Mask/LRR & Offset/Basis & FIL  & RE    & LT     & LL    & SF    & LF    & Avg \\
    \midrule
     2) & Mask & Offset & & 0.73(+151.72\%) & 1.01(+87.04\%) & 1.03(+58.46\%) & 0.92(+50.82\%) & 0.70(+70.73\%) & 0.88(+76.00\%) \\
     3) & Mask & Basis & & 0.53(+82.76\%) & 1.05(+94.44\%) & 1.04(+60.00\%) & 0.95(+55.74\%) & 0.56(+36.59\%) & 0.83(+66.00\%) \\
     4) & Mask & Basis & \checkmark & 0.45(+55.17\%) & 1.02(+88.89\%) & 0.93(+43.08\%) & 0.96(+57.38\%) & 0.50(+21.95\%) & 0.77(+54.00\%) \\
     5) & LRR & Basis & & 0.37(+27.59\%) & 0.69(+27.78\%) & 0.75(+15.38\%) & 0.75(+22.95\%) & 0.45(+9.76\%) & 0.60(+20.00\%)  \\
    \midrule
     6)& LRR & Basis & \checkmark & \textcolor{red}{0.29(+0.00\%)} & \textcolor{red}{0.54(+0.00\%)} & \textcolor{red}{0.65(+0.00\%)} & \textcolor{red}{ 0.61(+0.00\%)} & \textcolor{red}{0.41(+0.00\%)} & \textcolor{red}{0.50(+0.00\%)} \\
    \bottomrule
    \end{tabular}
    }\vspace{2pt}
    \caption{The homography representation solution is chosen in Offset and Basis (homography flow), as well as the outlier rejection module is chosen as Mask~\cite{zhang2020content} or LRR. The FIL as an optional promotion to the effect of the model.}\vspace{-3mm}
    \label{tab:ablation}
\end{table*}%
\begin{table}[t]
  \centering
  \small
  \resizebox{1\linewidth}{!}{
    \begin{tabular}{
    >{\arraybackslash}p{0.8cm}
    >{\centering\arraybackslash}p{1cm}
    >{\centering\arraybackslash}p{1.1cm}
    >{\centering\arraybackslash}p{1.1cm}
    >{\centering\arraybackslash}p{1.1cm}
    >{\centering\arraybackslash}p{1.3cm}
    }
    \toprule
     &0 layer &1 layer &2 layers &3 layers &3 layers* \\ %
    \midrule
     FIL & 0.00 & 0.14 & 0.31 & 0.35 & 0.11 \\
     MSE & 0.76 & 0.69 & 0.61 & 0.60 & 0.50 \\
    \bottomrule
    \end{tabular}
    }\vspace{3pt}
    \caption{Comparison of point matching errors when $f(\cdot)$ contains a different number of convolution layers. 0 layer means network without $f(\cdot)$. * indicates FIL is involved in training.}
    \label{tab:FIL}\vspace{-4mm}
\end{table}%

\subsection{Ablation Studies}\label{sec:ablation}\vspace{-1mm}
We verify the effectiveness of all three contributions by ablation studies, and report the results in Table~\ref{tab:ablation}. \vspace{-5mm}
\paragraph{Homography flow vs. offsets.}
The 2nd row of Table~\ref{tab:ablation} shows the result of CA-Unsupervised~\cite{zhang2020content} that adopts offset representation with mask outlier removal. We replace its regression target from the offset to the weights of our homography flow bases, and observe a lower average error and comparable errors in LT, LL and SF scenes being achieved in Row 3. It demonstrates that even for a network not specially designed, the homography flow representation outperforms the old offset solution.\vspace{-5mm}
\paragraph{LRR blocks.}
We modify the network in Row 3 to our structure (Row 5) by removing the mask module and inserting LRR blocks into the homography estimator $h(\cdot)$. By this replacement, the combination of LRR blocks and homography flow produces a notable advantage, reducing the error by at least about \textbf{20\%} in all scenarios. A reasonable explanation is that the LRR blocks encourage the utilization of low-rank features for homography estimation, which is conducive to features extraction and outlier rejection.\vspace{-5mm}
\paragraph{FIL for warp-equivariance.}
We also verify the effectiveness of FIL in two structures, \ie the modified CA-Unsupervised~\cite{zhang2020content} one as in Row 3 and ours as in Row 5, in Table~\ref{tab:ablation}. By comparing Row 3 with 4, and Row 5 with 6, we can observe that errors are reduced from 0.83 to 0.77 ($-8\%$) and from 0.60 to 0.50 ($-17\%$). Especially in the scenes LL and LF, the error reduction is more significant.

To further investigate how FIL improves the optimization, we conduct another experiment by modifying the number of convolutional layers in $f(\cdot)$, from 0 to 3. As seen in Table~\ref{tab:FIL}, from $f(\cdot)$ containing 0 to 2 layers without FIL involved in the optimization, the Mean Square Error (MSE) is gradually reduced while the FIL is increased, reflecting the warp-equivariance is damaged. Here ``0 layers'' means $f(\cdot)$ is removed and photometric loss is used instead. If we continue adding layer to 3, the MSE cannot be decreased showing that the network cannot be consistently optimized, while FIL keeps increasing. If we add FIL to the optimization goal, we can see the MSE is reduced from 0.6 to 0.5 and FIL is decreased significantly from 0.35 to 0.11. This phenomenon reflects that with warp-equivariance preserved, the optimization becomes more stable so that a higher performance can be achieved.

\subsection{Generalization}
\begin{figure}[!t]
\centering
   \includegraphics[width=1\linewidth]{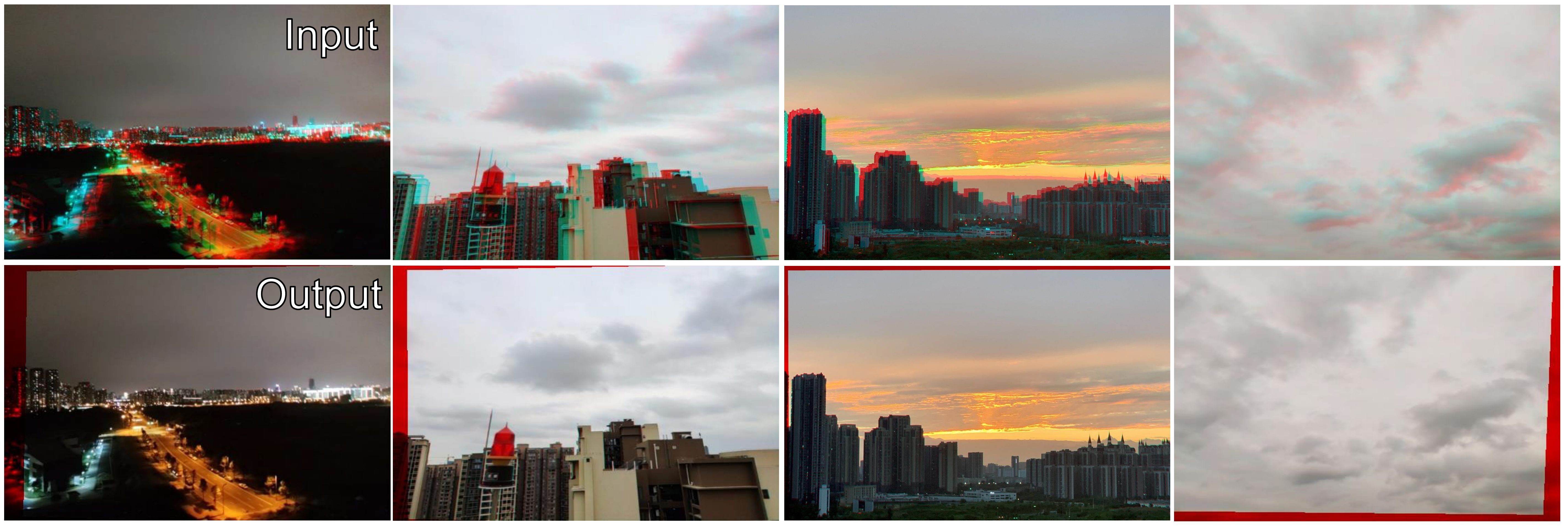}
   \caption{Photos taken by mobile phone.}
\label{fig:img_real}\vspace{-5mm}
\end{figure}

\vspace{-1mm}
As the fixed bases are obtained via mathematical derivation, its generation is independent of images, meaning that unseen images could be also covered as long as they are in small-baseline scenes. \fig~\ref{fig:img_real} depicts the alignment results of unseen photos taken with a mobile phone in scenes of low light and low texture, etc.

\subsection{Failure cases}\vspace{-1mm}
The predefined flow bases is friendly to small-baseline scenarios. By contrast, it may introduce error when applied to the large baseline cases as bases $h_7$ and $h_8$ are a linear approximate representation of small range perspective transformation in the Cartesian coordinates.
\vspace{-2mm}

\section{Conclusion}\vspace{-1mm}

We have presented a new deep solution for homography estimation, involving 3 new components to improve the performance of previous methods: a new representation called homography flow, a LRR block to reduce rank of features and a feature identity loss to stabilize the optimization process. Extensive experiments demonstrate the effectiveness of all the newly introduced components and the superior performance over the previous methods. Notwithstanding, our method has its limitations including that it may fail in large-baseline scenes, its single homography output may be insufficient for a real scene, and its fixed bases may limit its wider applications. We consider to extend the idea to mesh-based multi-homographies and explore the superiority of learned bases as our future work.

\vspace{-2mm}

\section*{Acknowledgement}
This research was supported in part by National Key R\&D Plan of the Ministry of Science and Technology (Project No. 2020AAA0104400), in part by National Natural Science Foundation of China (NSFC) under grants No.61872067 and No.61720106004, in part by Research Programs of Science and Technology in Sichuan Province under grant No.2019YFH0016.


{\small
\bibliographystyle{ieee_fullname}
\bibliography{egbib}
}

\end{document}